# Machine Learning for Paper Grammage Prediction Based on Sensor Measurements in Paper Mills

Hosny Abbas

*A Senior Engineer at Quena Paper Company, Qena, Qus, Egypt*
*A Ph.D. in Computers and Systems Engineering, Assiut University, 2017, Egypt*
Email: hosnyabbas@aun.edu.eg

*Abstract-* **Automation is at the core of modern industry. It aims to increase production rates, decrease production costs, and reduce human intervention in order to avoid human mistakes and time delays during manufacturing. On the other hand, human assistance is usually required to customize products and reconfigure control systems through a special process interface called Human Machine Interface (HMI). Machine Learning (ML) algorithms can effectively be used to resolve this tradeoff between full automation and human assistance. This paper provides an example of the industrial application of ML algorithms to help human operators save their mental effort and avoid time delays and unintended mistakes for the sake of high production rates. Based on real-time sensor measurements, several ML algorithms have been tried to classify paper rolls according to paper grammage in a white paper mill. The performance evaluation shows that the AdaBoost algorithm is the best ML algorithm for this application with classification accuracy (CA), precision, and recall of 97.1%. The generalization of the proposed approach for achieving cost-effective mills' construction will be the subject of our future research.**

*Keywords- industrial informatics, automation, machine learning, paper grammage classification, paper mills.*

## I. INTRODUCTION

Nowadays, there is a very active orientation towards using the Artificial Intelligence (AI) [1][2] paradigm in industrial applications and factory floor. ML represents a mainstream AI research area and there are many industrial applications designed and implemented based on ML techniques and algorithms. Examples of ML-based industrial applications are sales prediction, predictive maintenance, yield optimization, asset management and supply chain management, production optimization, product defects detection, products classification, just to mention a few. Typically, classification, regression, and clustering are the common activities of ML in industrial applications [3].

Modern automation systems are designed and deployed with a multi-layered architecture [4]. Field devices are situated in the bottom layer; control systems are situated in the middle layer, and supervisory stations or Human Machine Interfaces (HMIs) [5][6] are situated in the top layer. The top layer can be engineered to enable an internet access to the industrial process for supporting a remote operator or an expert engineer [7]. Because of the real-time constraints imposed on the two lower layers (the bottom and middle layers) and the necessity to work in a time-based deterministic fashion, in addition to the limited computational resources, ML techniques are rarely applied there. The top layer where supervisory assets are situated represents the common place for applying and utilizing ML techniques. In this layer, computers with higher computational power and storage capacity are found, that is why most of ML research applications are done in the top layer of the automation system. The HMI station is configured to connect to control systems such PLC (Programed Logic Controller) [8] or DCS (Distributed Control System) [9] with proprietary or standard communication protocols (i.e. Open Process Control (OPC) [10]), so that process data is available to the HMI application and can be used for predictive modeling [11].

In automatically controlled industrial processes, the job of the human operator is limited to supervisory and monitoring with less intervention with the process. From the other hand, he must give attention to the changes he perceives in the process through his HMI and in some cases he have to intervene to reconfigure or tune one or more process variables to keep the process in a stable and controlled state. Unfortunately, this manual behavior makes the control process vulnerable to human mistakes and time delays. With ML techniques, the high automation feature of control processes can be enhanced. ML algorithms can be used to predict the best process variable setpoint based on the values of some other process variables. Further, ML algorithms can also be used to offer the human operator the most appropriate actions he should take case of emergent situations. With ML techniques, the time when a robot operator replaces the human operator is approaching for the sake of saving time and avoiding human mistakes.

In this research, we try to use the most accurate ML algorithm to predict the paper roll Grammage (Basis Weight in $g/m^2$) based on real-time sensor measurements. There are one sensor for measuring the paper roll's diameter in (mm), another one for measuring its width in (mm), and a third one for measuring its Wight in (Kg).

Roughly speaking, the proposed ML approach comprises the following tasks:
1. Collecting process data and constructing the dataset.
2. Refining the dataset by removing non-logical entries and outliers.
3. Using a software tool such as Orange machine learning framework [1] to test the dataset and applying it to several Machine learning algorithms.
4. Selecting the most accurate algorithm based on the evaluation results obtained from Orange.
5. Implementing the selected model in python programming language using its ML library, *scikit-learn,* and saving the trained model for future predictions.
6. Developing a simple HMI application in python and configure a real connection to the PLC.
7. Test and verify the application run-ability with real-time sensor measurements.
8. Developing a complete HMI application in python programming language to take benefit of its several and valuable ML libraries.

The remaining of this research paper is organized as follows. Section II states the research problem in some details. Section III explores the related work to the application of ML algorithms in the industrial domain and factory floor. Section IV presents the proposed solution to the problem based on ML algorithms, including dataset collection and preparation, ML algorithm selection, implementation, evaluation, deployment and testing. Section V provides a future conceptual generalization of the proposed solution for minimizing the total cost of mills' construction and maintenance. Finally, Section VI concludes the paper and highlights future intentions.

## II. PROBLEM STATEMENT

Typically, a paper mill comprises three main stations: Paper machine, Winder station, and Wrapping station. The Paper machine produces paper with particular grammage in gsm (gram per square meter). The typical grammage classes in our paper mill are 48 gsm, 50 gsm, 58 gsm, 60 gsm, 68 gsm, 70 gsm. The Winder station takes a paper spool that is about 6 m width as it's input and transfers is to customized paper rolls with particular diameter and width. And, the Wrapping station is used for packaging purposes where the white paper roll is packaged with 2 or 3 layers of board paper (see Figure 1). Our ML industrial application will be carried out in the wrapping station. After a paper roll enters this station, its properties such as its diameter, width, and weight are measured using physical sensors. On the other hand the paper grammage in gsm or gram per meter square is not measured physically (there is no physical sensor for measuring it in wrapping station) and therefore the wrapping operator have to communicate from time to time with the process operator to get this information. The wrapping operator needs the paper grammage because he must write it manually to the wrapping station HMI application to be printed on the roll label for efficient storage of paper rolls. That causes time delays in the wrapping station and makes it vulnerable to operator mistakes.

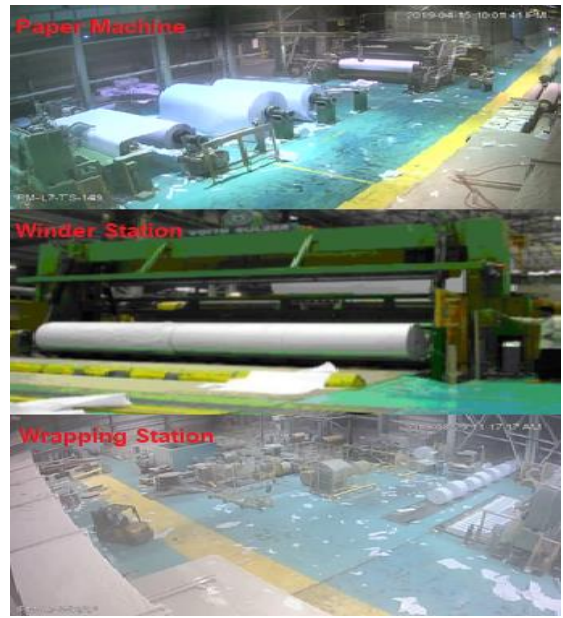

Fig. 1. A Paper Mill

In this research, we provide a solution to this problem by using an appropriate machine learning algorithm to automatically predict the paper grammage based on sensor measurements (roll properties such as diameter, width, and weight). Figure 2 demonstrates the process workflow.

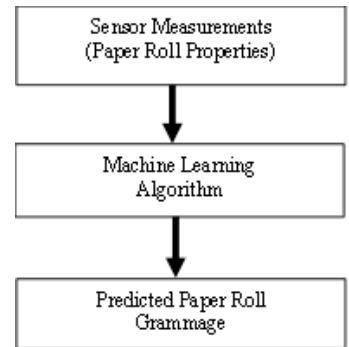

Fig. 2. Process Workflow

As shown in the figure, the input to the ML algorithm represent the real-time sensor measurements for each paper roll (diameter, width, and weight), and the output represents the predicted non-measurable process variable (i.e., paper grammage). The used dataset has been collected in the duration of one month production (about 10000 instances). TABLE 1 provides a sample segment of this dataset (just ten instances). As shown, it's a multi-class classification problem and an appropriate strong classifier with high accuracy and low variance is required.

TABLE 1: SAMPLE IN/OUT INSTANCES FROM THE APPLICATION DATASET

| Measured Roll Properties X | | | Output Variable y |
|---|---|---|---|
| Diameter (mm) | Width (mm) | Weight (Kg) | Grammage (gsm) |
| 900 | 820 | 444 | 70 |
| 900 | 1000 | 539 | 70 |
| 1000 | 810 | 576 | 70 |
| 1000 | 820 | 589 | 68 |
| 1000 | 820 | 589 | 70 |
| 1000 | 840 | 612 | 70 |
| 1000 | 1000 | 543 | 54 |
| 1000 | 1000 | 708 | 68 |
| 1000 | 1000 | 726 | 70 |
| 1000 | 860 | 603 | 70 |

## III. RELATED WORK

AI and its mainstream research branch 'ML' is currently playing a major role in the industrial domain to engineer and manage the increasing complexity in manufacturing processes and the national critical infrastructure utilities (power grids, water transportation, oil and gas, etc.). These systems produce a huge amount of process data which represent a plentiful and precious source of critical information. The need for professional data mining researchers who can apply variety of AI and ML techniques to get useful information which can be used to manage and control the complex modern industrial systems, has become a necessity.

Monostori [12] discussed the possibility to use AI and ML techniques for managing complexity, dynamic changes and uncertainties in manufacturing. He concluded that AI and ML can be effectively utilized to solve, within certain limits, unprecedented, unforeseen problems on the basis of even incomplete and imprecise information. Monostori's conclusions are compatible with the conclusions of Lu [13] who highlighted our urgent need for new computer technologies that can't only generate, record, and retrieve information, but also understand and synthesize information into knowledge and represent this knowledge properly to support decision making in modern complex real-life applications.

Table 2 provides some recent researched ML-based industrial applications. ML techniques have been utilized for solving challenging problems in factory floor such as determining the root causes of failures and defects in near-real-time. Traditionally, these non-trivial tasks are carried out by the human operators who have to analyze and handle a large amount of process data in the form of alarms, trends, and numerical values through the automation system HMI. As shown in the table, there are a variety of industrial applications and a variety of ML techniques, conventional and advanced. One of the emerging ML techniques that is considered as a revolution in computer learning approaches is Deep Learning (also called Hierarchical Learning). DL can be defined as a set of algorithms in machine learning that attempt to learn in multiple levels, corresponding to different levels of abstraction. It typically uses artificial neural networks. With DL higher-level features and concepts are defined in terms of lower-level ones, and such a hierarchy of features is called a deep architecture [31].

TABLE 2: SAMPLE MACHINE LEARNING INDUSTRIAL APPLICATIONS

| Authors | Application | Algorithm |
|---|---|---|
| Rutqvist et al. [14] | Smart Waste Management Systems | Random Forests |
| Lyutov et al. [15] | Managing customer requirements and specifications for an automotive industry company | Several conventional machine learning algorithms. |
| Golkarnarenji et al. [16] | Prediction of carbon fiber mechanical properties. | Support Vector Regression (SVR) and Artificial Neural Network (ANN). |
| Li, X [17] | Rolling bearing fault diagnosis. | Support Vector Machines. |
| Wu, C. [18] | Intelligent fault diagnosis of rotating machinery | Convolutional Neural Network (CNN). |
| Chen, J. [19] | Evaluate the welding quality of high-power disk laser welding. | Support Vector Machines. |
| Susto, G. [20] | Predictive Maintenance | SVM and kNN |
| Backus, P.[21] | An estimate of cycle time for a product in a factory. | Clustering, Trees, kNN |
| Li, L. [22] | Efficient Manufacture Inspection System | Deep Learning (CNN) |
| Priore [23] | Dynamic jobs scheduling in flexible manufacturing systems. | Inductive learning, backpropagation neural networks, and case-based reasoning (CBR). |
| Othman [24] | Voltage stability analysis of photovoltaic generation (PVG) system. Tuning PI controllers. | Genetic Algorithm (GA) technique |
| Mittal [25], Lu, W. et al. [26], Rivas-Echeverria [27], May, Z.,[28], Murthy [29] | Tuning of PID Parameters | Artificial Neural Networks (ANN) |
| Fu, Y. et al. [30], | Machines monitoring and fault diagnosis | Convolutional Neural Networks |
| Keynia [31] | Electricity price forecasting | Convolutional Neural Networks |
| Şeker et al. [32] | Condition monitoring in nuclear power plant and rotating machinery | Recurrent Neural Nets |

A key reason for the present interest and success of the DL approach is the massive improvements in computational horsepower brought about by Graphical Processor Units (GPUs) [32]. Another important reason to DL popularity is the great advances achieved in the Big

Data research area [33] which enhanced the ability to collect, store, and operate over large amounts of data. The ability to process large numbers of features makes deep learning very powerful when dealing with unstructured data. However, deep learning algorithms are not adequate for simple problems with less complexity because they require access to a vast amount of data to be effective. In other words, DL is not adequate for applications with simple datasets with few features and instances. The DL technique has been adopted in several fields such as computer vision, natural language processing, cyber-physical systems, and of course the industrial domain. Some DL applications are also provided in Table 2. The DL approach uses several algorithms such as Convolutional Neural Networks (CNNs) [34], Recurrent Neural Networks (RNNs) [35], Generative Adversarial Networks [36], and Feedforward networks [37]. The research presented in this paper adopts conventional ML algorithms because of the small number of features used in the application dataset. In future, we intend to use the DL technique for paper defects detection through a smart Web Inspection System [40].

## IV. THE PROPOSED ML APPROACH

### A. Dataset Collection And Preparation

For one month of hard work, we were collecting and constructing the application dataset. Sample instances of the dataset were provided above in Table 1 (Section II). Each entry in the dataset has three input features (roll diameter, roll width, and roll weight) which are measured by three physical sensors, and one target variable (roll grammage) which is entered manually by the operator through his HMI. To refine our dataset, we used Orange software framework [41] which is a data mining framework for data analysis through python scripting and visual programming. Orange performs simple data analysis with clever data visualization. It supports interactive data exploration for rapid qualitative analysis with clean visualizations. The developer can place widgets on the canvas, connect them, load his datasets and harvest the insight. Figure 3 presents the initial Orange workflow for visualizing and refining our dataset.

We used the FreeViz (free visualization) widget to visualize the initial dataset. As shown in Figure 4, the dataset has 6 classes and there are some outliers. Because outliers can impact the performance of some ML classifiers and lower its accuracy, we used the Outliers Orange widget to remove the outliers and Figure 5 visualizes the dataset after excluding the outliers. The main cause of existence of outliers is sensors eventual malfunction, in addition to the human operator unintended mistakes. The new dataset was then entered as an input to an Orange DataSampler widget to divide it into training set (70%) and testing set (30%). Table 3 provides the dataset preparation results. The training and testing sets will be the inputs of several ML learners.

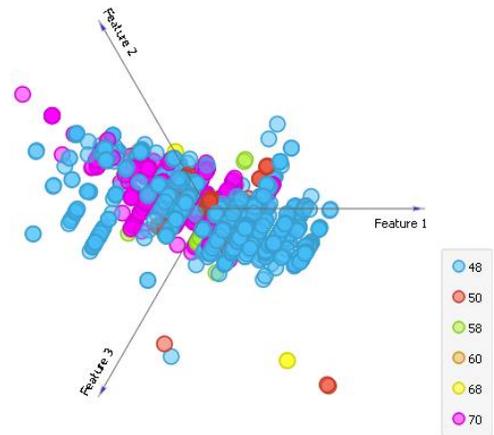

Fig. 4. Dataset Free Visualization before excluding outliers

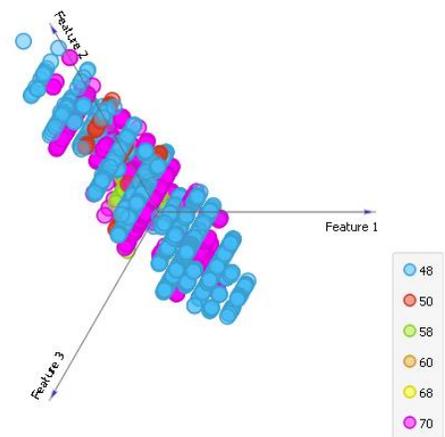

Fig. 5. Dataset Free Visualization after excluding outliers

TABLE 3: DATASET PREPARATION

| Total | Outliers | Inliers | Train Set | Test set |
|-------|----------|---------|-----------|----------|
| 9589  | 1918     | 7671    | 5370      | 2301     |

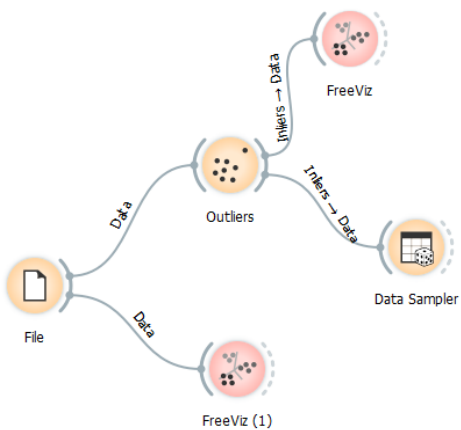

Fig. 3. Dataset Analysis with Orange Framework

## B. ML Algorithm Selection

We tested our dataset with nine ML classification algorithms {Neural Networks, Naïve Bayes, Stochastic Gradient Descent or SGD, Logistic Regression, Support Vector Machines or SVM, kNN, Tree, Random Forests, and AdaBoost). The complete Orange workflow is shown in Figure 6 and its evaluation results are provided in Figure 7. As shown in the evaluation results, many classifiers give high *Classification Accuracy (CA)*. The best of them is the AdaBoost algorithm which gives CA equal to 97.1% and the next is Random Forest which gives CA equal to 96.8%. It is clear from the evaluation results that the decision-tree based ML algorithms are more accurate in their predictions. That is related to the nature of our dataset.

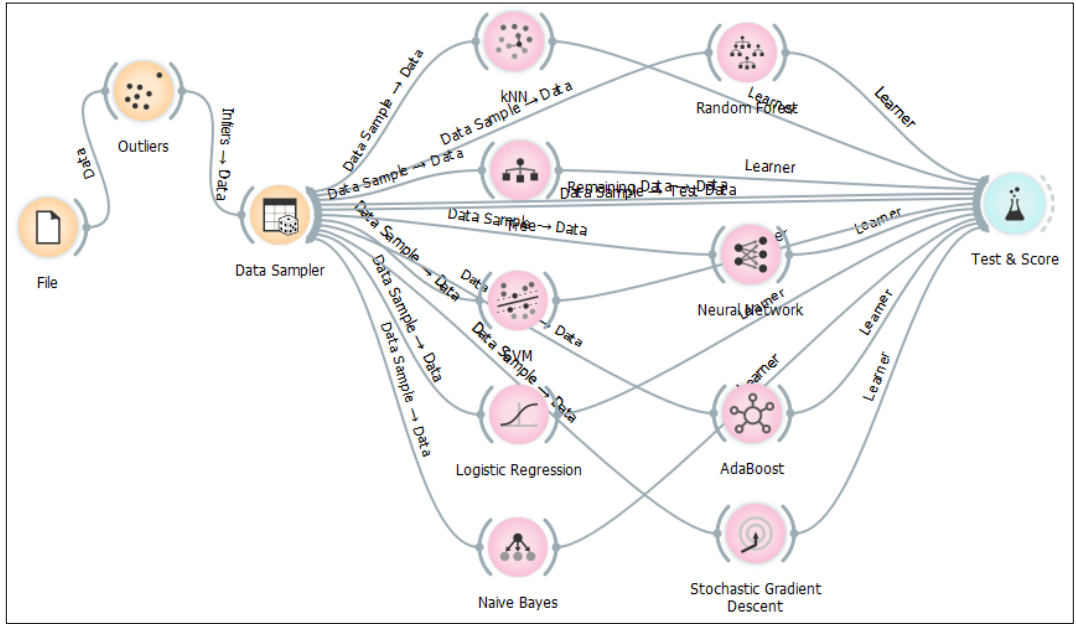

Fig. 6. The Designed Orange's Workflow with several ML Algorithms have been Applied on the dataset

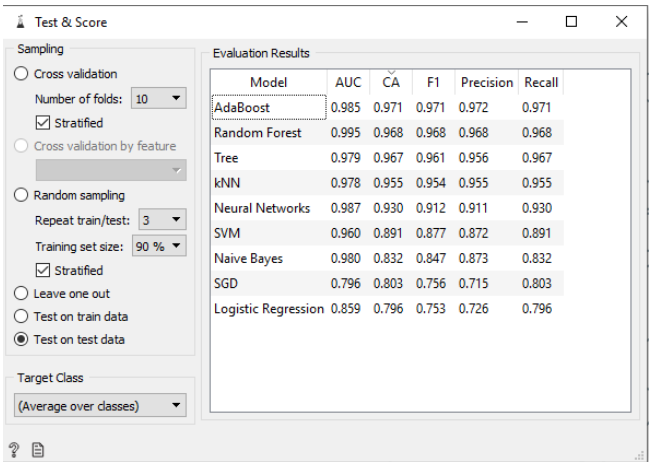

Fig. 7. Performance Comparison between the applied ML algorithms

Thanks to the used Orange data mining environment, we were able to test our application dataset easily, quickly and effectively. Now, we understand that any decision tree based ML algorithm can give us high CA. We selected the highest CA value classifier which is the AdaBoost algorithm. AdaBoost [42], short for Adaptive Boosting, is a machine learning meta-algorithm formulated by Yoav Freund and Robert Schapire [43], who won the 2003 Gödel Prize for their work. It can be used in conjunction with many other types of learning algorithms to improve performance. With AdaBoost as a meta-model, the output of other weak learners is combined into a weighted sum that represents the final output of the boosted classifier. AdaBoost is adaptive in the sense that subsequent weak learners are tweaked in favor of those instances misclassified by previous classifiers. Further, the AdaBoost algorithm is sensitive to noisy data and outliers. In some problems it can be less susceptible to the overfitting problem than other learning algorithms. The individual learners can be weak, but as long as the performance of each one is slightly better than random guessing, the final model can be proven to converge to a strong learner. AdaBoost makes predictions by applying multiple decision trees to every sample and combining the predictions made by individual trees. However, rather than taking the average of the predictions made by each decision tree in the forest (or majority in the case of classification), in the AdaBoost algorithm, every decision tree contributes a varying amount to the final prediction.

## C. Implementation, Deployment And Testing

In this stage we decided to use the AdaBoost classification model which gave us the highest accuracy,

precision, and recall (97.1%). We will re-implement the AdaBoost model with python programming language and using the *scikit-learn* library which supports most of ML algorithms. Because the random forest classifier has the highest accuracy (96.8%) next to the AdaBoost classifier (see Figure 7), we used it as its base classifier. The next code segment demonstrates the model implementation in python, see how to create, train, and test the model. The output of the code execution prints the accuracy of the created AdaBoost model (97.4%); it is very close to with the accuracy of the Orange evaluation results (97.1%). We saved the created AdaBoost model in a pickle file (*.pkcls) so that we can use it after reloading in other python applications.

```
# Load libraries
from sklearn.ensemble import AdaBoostClassifier
from sklearn.model_selection import train_test_split
from sklearn import metrics
import numpy as np
from sklearn.ensemble import RandomForestClassifier
import pickle
from warnings import simplefilter

# ignore all future warnings
simplefilter(action='ignore', category=FutureWarning)
# Loading Dataset
fileName='d46-Refined3.txt'
a = np.loadtxt(fileName)
X = a[:,:3]
y = a[:,3]
# 80% training and 20% test
X_train, X_test, y_train, y_test = train_test_split(X, y, test_size=0.2)
model= AdaBoostClassifier(RandomForestClassifier(max_depth=1),
            n_estimators=10,learning_rate=1,algorithm
                    ='SAMME.R',random_state=0)
print('Train Adaboost Classifer...')
model =model.fit(X_train, y_train)
print('predict.....')
y_pred = model.predict(X_test)
print("Accuracy:", metrics.accuracy_score(y_test, y_pred))
# Save the model for future predictions
pickle.dump(model, open('AdaBoost_model.sav', 'wb'))

Train Adaboost Classifer...
predict.....
Accuracy: 0.9741275167785235
```

We developed an HMI process interface in python. The developed HMI was prepared to connect to a real PLC system which controls the wrapping station. The model of the wrapping station PLC is Siemens s7-400. To connect with the PLC through python, the OPC process communication protocol [44] was used. OPC was originally stands for OLE (Object linking and Embedding) for Process Control. It delivers connectivity and interoperability benefits to measurement and automation systems in the same way that standard printer drivers deliver connectivity and interoperability benefits to word processing. To enable our python HMI to communicate with the OPC protocol which is a COM (Component Object Model) based API (Application Programming Interface), a library called OpenOPC [45] should be used. OpenOPC for Python is a free, open source OPC (OLE for Process Control) toolkit designed for use with the popular Python programming language. The library is simple to learn and easy to remember. Figure 8 presents the confusion matrix to describe the performance of the implemented AdaBoost classification model with Random Forest as its base classifier.

|  | Predicted | | | | | | |
|---|---|---|---|---|---|---|---|
|  | 48 | 50 | 58 | 60 | 68 | 70 | Σ |
| 48 | 97.7 % | 3.7 % | 5.1 % | 0.0 % | 0.0 % | 0.7 % | 2141 |
| 50 | 0.6 % | 92.6 % | 1.1 % | 0.0 % | 0.0 % | 0.0 % | 316 |
| 58 | 1.2 % | 3.1 % | 90.3 % | 0.0 % | 0.0 % | 0.6 % | 209 |
| 60 | 0.0 % | 0.0 % | 0.0 % | 100.0 % | 0.0 % | 0.1 % | 4 |
| 68 | 0.0 % | 0.0 % | 0.0 % | 0.0 % | 60.9 % | 0.7 % | 57 |
| 70 | 0.6 % | 0.6 % | 3.4 % | 0.0 % | 39.1 % | 98.0 % | 2643 |
| Σ | 2152 | 326 | 175 | 2 | 64 | 2651 | 5370 |

Fig.8. Performance Comparison between the applied ML algorithms

The next python code segment gives a simple example of how to load the trained and saved AdaBoost model and use it to predict the grammage of the current paper roll by establishing a connection to the Siemens PLC station using the OPC communication protocol and reading the paper roll properties (diameter, width, and weight) then entering this data to the ML model. The code segment also reads the actual roll grammage which was entered by the operator through the HMI. We asked the operator to continue entering manually the actual paper grammage for comparison with the predicted value. As shown in the code output, the predicted class exactly equals the actual class. The complete developed HMI application for wrapping station is shown in Figure 9. The complete HMI was partially developed and tested in run-time. It is still under development in order to add some other machine learning functions such as predicting the quality of each paper roll according to the specifications of the mill quality LAB, instead of manual entering by the operator.

## V. FUTURE CONCEPTUAL GENERALIZATION

A key goal of present and future manufacturers is to produce high quality products at minimum cost. AI and ML represent the promising technologies to achieve this goal especially with the advances made in data management and computation technologies which give ML techniques the capacity to analyze very large amounts of data in real time. From a conceptual perspective, the proposed approach can be used to reduce the construction cost of industrial mills (i.e., paper mills).

```
# Load libraries
import pickle
import OpenOPC
import time
import pywintypes
import numpy as np
# load the saved model
model=pickle.load(open('AdaBoost_model.sav', 'rb'))
# Connect to the OPC server
pywintypes.datetime = pywintypes.TimeType
opc=OpenOPC.client()
opc.connect('OPC.SimaticNET.1')
# read sensor measurements
diameter,q,t=opc.read('S7:[S7_connection_1|VFD3|S7ONLINE|01.00,192.168.1.153,01.02,1]db1,w2')
width,q,t=opc.read('S7:[S7_connection_1|VFD3|S7ONLINE|01.00,192.168.1.153,01.02,1]db1,w4')
weight,q,t=opc.read('S7:[S7_connection_1|VFD3|S7ONLINE|01.00,192.168.1.153,01.02,1]db1,w6')
manual_grammage,q,t=opc.read('S7:[S7_connection_1|VFD3|S7ONLINE|01.00,192.168.1.153,01.02,1]db1,w12')
predicted_grammage=model.predict([[diameter,width,weight]])
print("diameter"," ", "width", " ", "weight", "predicted_grammage", " ",
                                          "manual_grammage")
print("  ",diameter,"   ", width, "   ", weight, "       ",
                        int(round(predicted_grammage[0])), "           ", manual_grammage)
```

| diameter | width | weight | predicted_grammage | manual_grammage |
|----------|-------|--------|--------------------|-----------------|
| 1000     | 820   | 564    | 70                 | 70              |

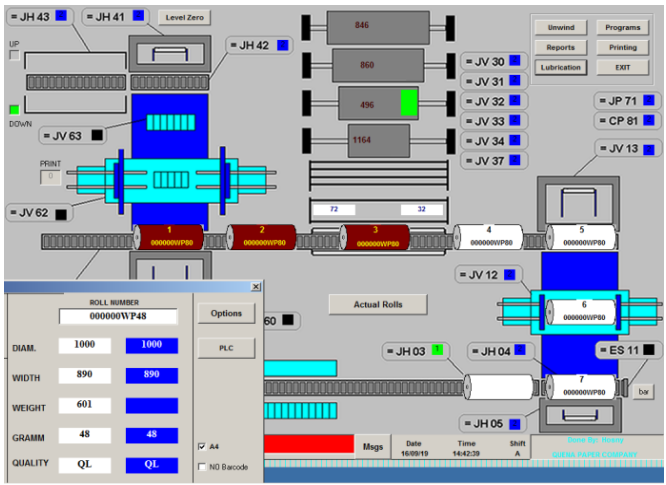

Fig.9. The Developed HMI for Wrapping Station

To explain, one of the crucial problems facing industrial production these days is the great complexity in manufacturing processes, which results in the installation of a large number of modern technological devices to measure a lot of product-related physical properties such as its diameter, width, weight, etc., and process-related physical quantities such as temperature, flow, level, pressure, and so forth. This large number of measuring devices represents a key factor in the increasing final cost of constructing a new industrial facility not only at the beginning of the construction process, but also during production because that necessitates the existence of large quantity of spare parts, that is in addition to the interruption of the production process from time to time in case of breakdowns in one or more of these devices. The proposed solution presented in this paper can be adopted to resolve this problem based on machine learning algorithms (either conventional or advanced algorithms). Machine learning algorithms can be used to reduce the number of physical measuring devices. In other words, based on the presence of few principal measuring devices, an appropriate machine learning algorithm can predict the value of other not-installed devices.

The future generalization of this research aims to provide a methodology for reducing the number of necessary physical measuring analog signal devices to reduce both of the total construction and maintenance costs. This approach is expected to be efficient especially with discrete production lines where there is a pipeline of similar products and their physical properties (i.e., weight, width, diameter, and so forth) are required to be measured. It can also be applied to continuously changing physical quantities (i.e. flow, temperature, level, and so forth) with carful design. To semi-formally describe the research problem and as demonstrated in Figure 10, suppose there is a product, (or more generally a process P that requires measuring a set of physical quantities or properties (D) which consists of a number of measuring instruments or devices ($d_1$, $d_2$, $d_3$, …,$d_p$). Using appropriate machine learning algorithms, it is aimed to reduce the number of these devices from p devices to p-k devices, where (p>k ≥1). In other words, based on the measured p-k devices values, k other measurements will be calculated (predicted) with machine learning algorithms. The challenging part of this generalization is the determination of the criteria adopted for selecting which process variables will be measured and which will be predicted. Actually, a systematic approach for device selection is required, that is left for future research.

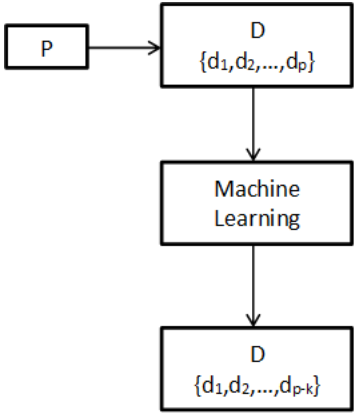

Fig.10. Using Machine Learning to reduce the required number of measuring devices in a complex industrial process.

## VI. CONCLUSIONS AND FUTURE WORK

Modern automation systems have many challenging problems for which the traditional model-driven engineering is limited. Data-driven methods used in ML research area can provide the required solutions to these challenging problems. Today's, there are a lot of successful real-life applications depend mainly on ML techniques in variety of domains. This paper provides an example of the industrial adoption of ML algorithms to help human operators save their mental effort and avoid time delays for the sake of high production rates. Based on real-time sensor measurements, several ML algorithms have been used to classify paper rolls according to their grammage. The performance evaluation shows that the AdaBoost algorithm is the best ML algorithm for this application with classification accuracy (CA), precision, and recall of 97.1%. A complete HMI application was partially developed in python programming language and it is under development and testing. Further, the generalization of the proposed approach for achieving cost-effective mills' construction will be investigated in future research. Finally, a smart web inspection system for detecting and classifying paper defects in paper mills using advanced machine learning algorithms (i.e., deep learning) is left as a future research too.